\pdfoutput=1
\documentclass[11pt]{article}

\usepackage{ACL2023}

\usepackage[utf8]{inputenc}
\usepackage[T1]{fontenc}
\usepackage{hyperref}
\usepackage{url}
\usepackage{booktabs}
\usepackage{amsfonts}
\usepackage{nicefrac}
\usepackage{microtype}
\usepackage{graphicx}
\usepackage{xcolor}
\usepackage{lipsum}
\usepackage{caption}
\usepackage{times}
\usepackage{latexsym}
\usepackage{graphicx}
\usepackage{booktabs}
\usepackage{threeparttable}
\usepackage{multicol}
\usepackage{caption}
\usepackage{subcaption}
\usepackage{float}
\usepackage{multirow} 
\usepackage{multicol} 
\usepackage{tabularx} 
\usepackage{framed}
\usepackage{enumitem}

\usepackage[T1]{fontenc}
\usepackage[utf8]{inputenc}

\usepackage{microtype}

\usepackage{inconsolata}

\title{Can We Trust LLMs? Mitigate Overconfidence Bias in LLMs through Knowledge Transfer}

\author{
  Haoyan Yang, Yixuan Wang, Xingyin Xu, Hanyuan Zhang, Yirong Bian \\
  Center for Data Science, New York University \\
  \texttt{\{hy2847,yw7872,xx943,hz1832,yb970\}@nyu.edu}
}

\newenvironment{promptbox}{
    \definecolor{shadecolor}{rgb}{0.93,0.93,0.93}
    \begin{shaded}
}{
    \end{shaded}
}

\begin{document}
\maketitle

\begin{abstract}
The study explores mitigating overconfidence bias in LLMs to improve their reliability. We introduce a knowledge transfer (KT) method utilizing chain of thoughts, where ``big'' LLMs impart knowledge to ``small'' LLMs via detailed, sequential reasoning paths. This method uses advanced reasoning of larger models to fine-tune smaller models, enabling them to produce more accurate predictions with calibrated confidence. Experimental evaluation using multiple-choice questions and sentiment analysis across diverse datasets demonstrated the KT method's superiority over the vanilla and question-answer pair (QA) fine-tuning methods. The most significant improvement in three key metrics, where the KT method outperformed the vanilla and QA methods by an average of 55.3\% and 43.1\%, respectively. These findings underscore the KT method's potential in enhancing model trustworthiness and accuracy, offering precise outputs with well-matched confidence levels across various contexts.
\end{abstract}
\setlength{\parskip}{0em}

\section{Introduction}
\begin{figure}[t]%
	\centering
    \small
	\includegraphics[width=0.48\textwidth]{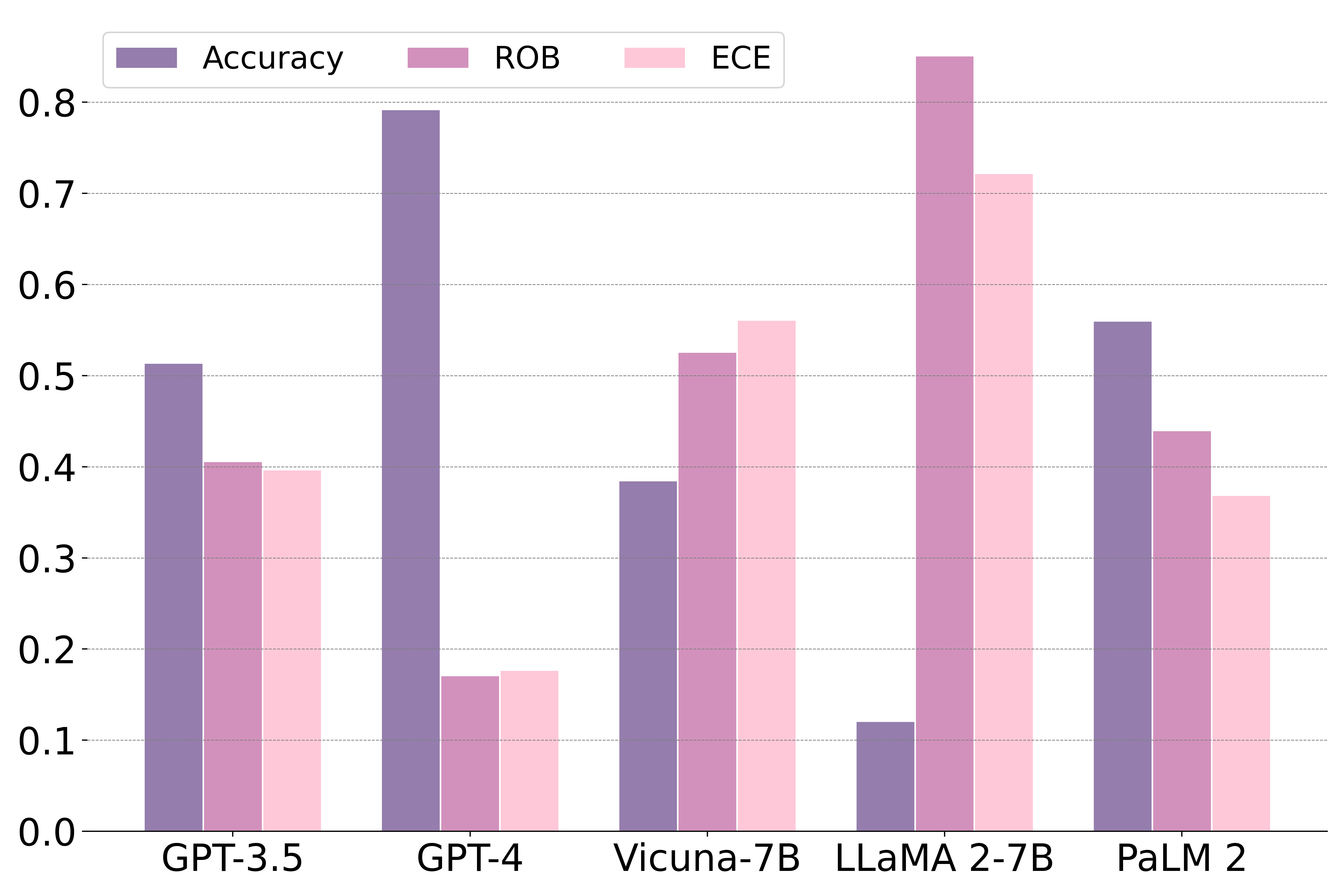}
	\caption{Evaluation of overconfidence in LLMs using the TruthfulQA dataset. Graphs indicate a pronounced discrepancy between confidence and correctness. Especially, like LLaMA 2 and Vicuna exhibit instances of high confidence in incorrect answers (high ROB), underscoring a widespread overconfidence issue in current LLMs.}\label{fig1}
\end{figure}

Large Language Models (LLMs), such as GPT-4 \cite{openai2023gpt4} and PaLM 2 \cite{anil2023palm}, provide benefits to solve real-life tasks. However, ``overconfidence bias'' in LLMs, where models yield overconfidence in their incorrect predictions, is a notable issue. As demonstrated by initial tests on the Truthfulqa \cite{lin2022truthfulqa} dataset in Figure \ref{fig1}, they reveal a significant overconfidence bias in several leading LLMs, especially Vicuna\footnote{Our experiments were conducted with Vicuna-7B-1.5 via \url{https://huggingface.co/lmsys/vicuna-7b-v1.5}.} and LLaMA 2 \cite{touvron2023llama}.

Mitigating overconfidence in models is vital for accuracy and trust in key sectors such as law and finance, to prevent costly errors and enhance reliability. Therefore, in this paper, we proposed a knowledge transfer (KT) method to address this issue rely on fine-tuning through Chain of Thoughts (CoT). 

Previous studies have improved LLMs' responses using the CoT strategy, enhancing reasoning transparency, with notable contributions from Wei et al. \cite{DBLP:journals/corr/abs-2201-11903} and Huang et al \cite{huang2022large}. However, these did not address LLMs' overconfidence bias. Our research aims to use CoT to tackle the issue.

\section{Related Work}
\textbf{Confidence level in LLMs}\indent The previous research highlighted limitations in logit-based confidence assessment  of language models \cite{wei2022mitigating}. Others pointed out that logits can cause overconfidence and do not adequately capture uncertainty \cite{Stephanie2022express}. Furthermore, they are not applicable to closed-source models like GPT-4. Alternatively, recent works \cite{liu2023trustworthy, xiong2023can, zhang2023towards} proposed using language models' verbalized confidence as a more effective measurement. We followed the ``confidence elicitation'' method by Miao et al. \cite{xiong2023can}, to better gauge their confidence levels.\\\\
\noindent\textbf{Stimulation of Reasoning Capabilities through CoT}\indent Wei et al. \cite{DBLP:journals/corr/abs-2201-11903} introduced the CoT prompting strategy to enhance the reasoning transparency of LLMs. This method provides structured insights into how these models process information. Following this, Shridhar et al. \cite{shridhar2023distilling} and Ma et al. \cite{ma2023sci} applied CoT in distilling step-by-step reasoning to fine-tune smaller models. Shridhar et al. focused on general reasoning, while Ma et al. concentrated on improving the mathematical.

\section{Methodology}
\begin{figure}[h]%
	\centering
    \small
	\includegraphics[width=0.48\textwidth]{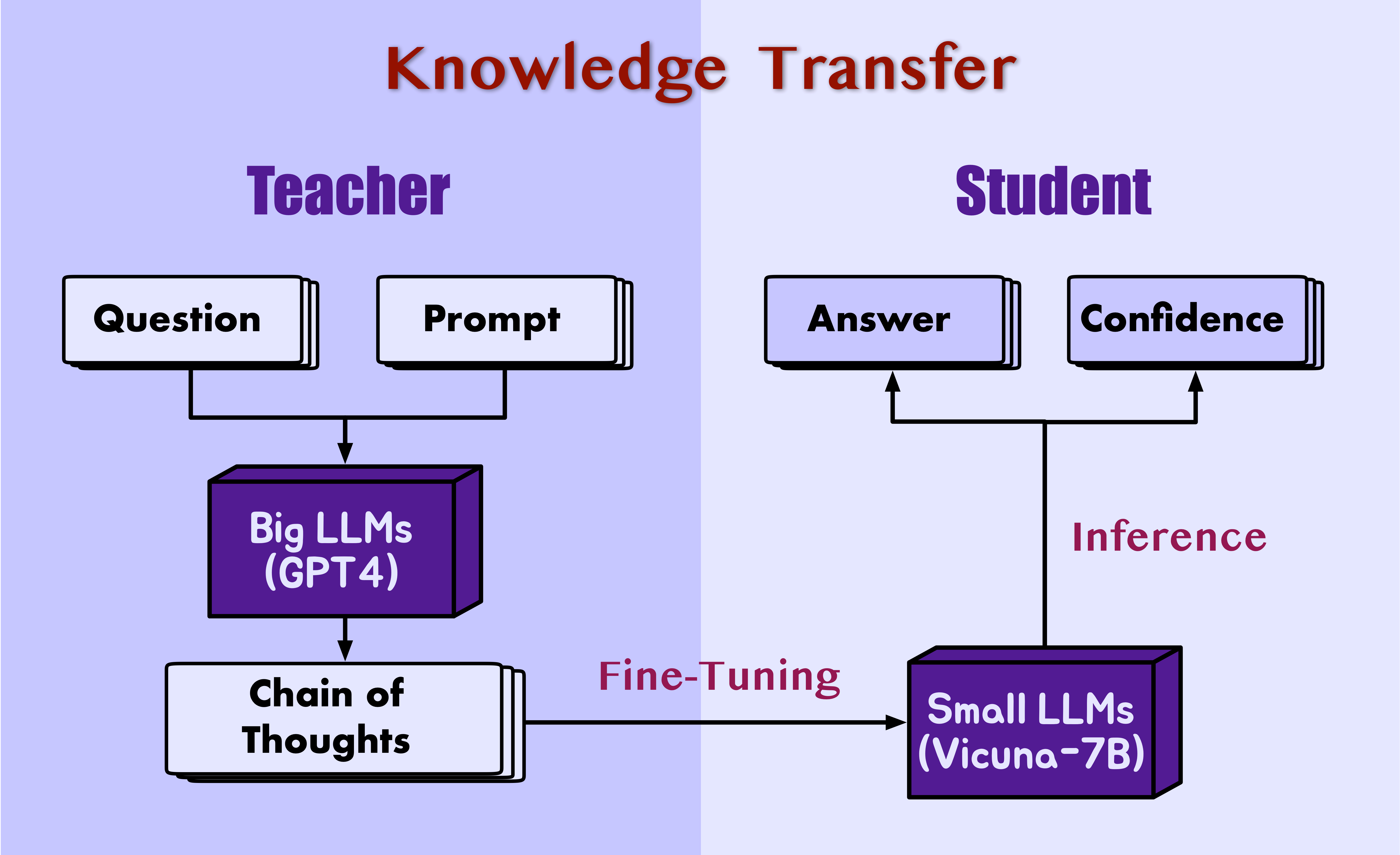}
	\caption{Schematic representation of the KT method between ``big'' and ``small'' LLMs. The diagram shows the role of big LLMs, such as GPT-4, as the ``Teacher'', where they generate detailed CoTs given a ``Question'' and a ``Prompt''. This output is then utilized to fine-tune small LLMs, exemplified by 'Vicuna-7B' as the ``Student''. The fine-tuned models then employ these insights to enhance their inference capabilities, yielding answers with quantifiable confidence levels.}\label{fig2}
\end{figure}

In Figure \ref{fig2}, we presented the KT method  where ``big'' LLMs like GPT-4 (as ``Teacher'') transfer knowledge via CoTs to fine-tune ``small'' LLMs like Vicuna-7b (as ``Student''), aiming to correct the smaller models' overconfidence bias.

\subsection{Knowledge Extraction From ``Big'' LLMs}
``Big'' LLMs generate CoTs from questions and prompts as follows. These CoTs include detailed and sequential reasoning pathways, voicing how the model arrives at its correct answers.

\begin{promptbox}
\noindent\textbf{Prompt:} Read the sentence, analyze the sentiment step by step, select the correct option and give option letter e.g. A or B as your answer. Use the following format to provide your answer and confidence level:

\noindent\textit{Explanation:} [insert step-by-step analysis here]

\noindent\textit{Answer and Confidence (0-100):} [Your answer, please only include the capital letter, e.g. B], [Your confidence level, please only include the numerical number, e.g. 80]

\noindent\textit{Note:} The confidence level indicates the degree of certainty you have about your answer and is represented as a percentage. For instance, if your confidence level is 80\%, it means you are 80\% certain that your answer is correct and there is a 20\% chance that it may be incorrect.

\noindent\textbf{Question:} A rollicking good time for the most part. \textit{Options:} A. Positive B. Negative.

\end{promptbox}

\subsection{Fine-tuning ``Small'' LLMs}
Once generated, CoTs as follows are combined with question to fine-tune the ``small'' LLMs. 
\begin{center}
    \fbox{
        \begin{minipage}{0.48\textwidth}
            <s> [INST] Question: a rollicking good time for the most part Options: A. Positive B. Negative [/INST] Explanation: The sentence speaks about a 'rollicking good time' 'Rollicking good time' refers to a lively, enjoyable, and amusing time The latter part of the sentence, 'for the most part' indicates that, for the majority of the time, the subject was having a jovial, good time This denotes a positive sentiment in the sentence, as it talks about a generally enjoyable time or experience </s>
        \end{minipage}}
    % \captionof{figure}{dasda}
\end{center}
The fine-tuning aims to integrate the knowledge and reasoning patterns of ``Big'' LLMs into ``small'' LLMs, allowing the smaller one to mimic the advanced reasoning capabilities of its larger counterpart. The loss function of fine-tuning can be mathematically described as
\begin{equation}
    L(\theta) = - \sum_{i=1}^{N} y_i \log(f_\theta(x_i))
\end{equation}
where $\theta$ represents the parameter of ``small'' models; $N$ is the number of text; $y_i$ is the true next token and $f_\theta$ is the probability distribution over possible next tokens predicted by ``small'' LLMs for the $i_{th}$ token $x_i$ from CoT text.

\subsection{Confidence-Calibrated Inference}
The fine-tuned ``small'' LLMs utilize their learned reasoning abilities. By employing the prompt outlined in Appendix \ref{appendix1}, models can generate predicted answers along with an associated confidence level, indicating the degree of certainty it has about the response.

\section{Experiments}

\subsection{Tasks and Datasets}
In our experimental evaluation, we focused on two tasks: multiple-choice questions and sentiment analysis. Table \ref{table1} demonstrates the datasets we used for each task and Appendix \ref{appendix2} shows the details.

\begin{table}[h]
\centering
\begin{threeparttable}
\small
\caption{Datasets used in multiple-choice questions and sentiment analysis task.}
\label{table1}
\begin{tabular}{p{1.5cm}p{5.3cm}}
\toprule
\textbf{Task} & \textbf{Dataset} \\
\midrule
Multiple-choice Questions & Truthfulqa\cite{lin2022truthfulqa}, McTest\cite{richardson-etal-2013-mctest}, RACE\cite{lai2017race}, ARC\cite{clark2018think}\\
Sentiment Analysis & SST2 \cite{wang2019glue}, Financial Phrasebank\tnote{1}, Twitter\cite{paws2019naacl}, GooglePlay\tnote{2}\\
\bottomrule
\end{tabular}
\begin{tablenotes}
\item[1] Our experiments will be conducted on Financial Phrasebank dataset via \url{https://huggingface.co/datasets/financial_phrasebank}.
\item[2] The GooglePlay dataset was constructed by our team, with all data sourced from \url{https://play.google.com/store/games?hl=en_US&gl=US}.
\end{tablenotes}
\end{threeparttable}
\end{table}

\subsection{Baselines}
We selected two open-sourced models which have serious overconfident bias as our baselines: LLaMA 2-7B \cite{touvron2023llama} and Vicuna-7B. By comparing three methods: zero-shot using the baseline model (\textbf{Vanilla}), fine-tuning with question-answer pairs (\textbf{QA}), and our \textbf{KT} method, we aim to validate the effectiveness of our proposed method in alleviating the issue of overconfidence bias in LLMs.

\subsection{Experimental Setting}
We use the QLoRA  \cite{dettmers2023qlora} approach for fine-tuning the baseline model, with the main parameter settings as shown in Table \ref{table2}.

{
\setlength{\tabcolsep}{13.5pt}
\begin{table}[h]
\centering
\small
\caption{Experimental hyperparameters settings during training and inference processes}
\label{table2}
\begin{tabular}{ccc}
\toprule
& \textbf{Training} & \textbf{Inference} \\
\midrule
\textbf{Hyperparameter} & \textbf{Value} & \textbf{Value} \\
\midrule
LoRA dim & 64 & \\
Alpha & 16 & \\
Dropout & 0.1 & \\
Epochs & 20 & \\
Batch size & 4 & \\
LR & 2e-4 & \\
Weight decay & 0.001 & \\
Optimizer & Adam & \\
Warmup & 0.03 & \\
Sample & & True \\
Temp. & & 0.7 \\
Top-p & & 0.95 \\
Top-k & & 5 \\
Max tokens & & 512 \\
\bottomrule
\end{tabular}
\end{table}
}

\subsection{Metrics}
\textbf{Accuracy (ACC):} Ratio of correct answers. \\
\textbf{Ratio of Overconfidence Bias (ROB):} Ratio of incorrect answer with a confidence level above 80\%. \\
\textbf{Expected Calibration Error (ECE):} Discrepancy between model's confidence levels and outcomes. 

\subsection{Comparative Results}
Upon examining the comparative results in Table \ref{table3}
, the KT method exhibits consistent improvements on most datasets.  Notably, within the LLaMA 2-7B model framework of Truthfulqa dataset, KT's accuracy substantially exceeds that of the Vanilla model by 64.4\% and outperforms the QA method by 47.8\%. Furthermore, KT reduces ROB by 61.6\% over Vanilla and 47.8\% over QA, decreasing ECE by 39.8\% and 33.8\% respectively. 

Additionally, KT's ECE metrics are 0.039 in the LLaMA-GooglePlay and 0.051 in the Vicuna-SST2 configurations, demonstrating the model's enhanced reliability with the KT method. 

Overall, this significant improvement not only highlights the effectiveness of KT in various contexts but also its capacity to make models more trustworthy.

\begin{table*}[t]
\centering % Center the table
\small
\caption{Comparative performance metrics of LLaMA 2-7B and Vicuna-7B models using Vanilla, QA, and KT approaches across various datasets.}
\label{table3}
\begin{tabularx}{\textwidth}{XXXX|XXX|XXX|XXX}

\hline
& \multicolumn{6}{c|}{\texttt{LLaMA 2-7B}} & \multicolumn{6}{c}{\texttt{Vicuna-7B}} \\
\hline

% Use the \multirow command to merge rows and \multicolumn to merge columns where necessary
& \multicolumn{3}{c|}{Truthfulqa} & \multicolumn{3}{c|}{SST2} & \multicolumn{3}{c|}{Truthfulqa} & \multicolumn{3}{c}{SST2} \\
\cline{2-13}
& Vanilla & QA & KT & Vanilla & QA & KT & Vanilla & QA & KT & Vanilla & QA & KT \\
\hline
ACC & 0.121 & 0.287 & \textbf{0.765} & 0.721 & 0.758 & \textbf{0.864}  & 0.385 & 0.494 & \textbf{0.701} & 0.698  & 0.684 & \textbf{0.837} \\
ROB & 0.851 & 0.713 &  \textbf{0.235} & 0.247 & 0.219 & \textbf{0.085} & 0.526 & 0.418 & \textbf{0.152} & 0.297  & 0.311 & \textbf{0.132} \\
ECE & 0.722 & 0.663 &  \textbf{0.324} & 0.206 & 0.198 & \textbf{0.099}  & 0.561 & 0.412 & \textbf{0.135} & 0.231  & 0.306 & \textbf{0.051} \\
\hline
& \multicolumn{3}{c|}{McTest} & \multicolumn{3}{c|}{Financial Phrasebank} & \multicolumn{3}{c|}{McTest} & \multicolumn{3}{c}{Financial Phrasebank} \\
\cline{2-13}
& Vanilla & QA & KT & Vanilla & QA & KT & Vanilla & QA & KT & Vanilla & QA & KT \\
\hline
ACC & 0.533 & 0.652 & \textbf{0.663} & 0.396 & 0.428 & \textbf{0.466}  & 0.683 & 0.645 & \textbf{0.756} & 0.325  & 0.434 & \textbf{0.834} \\
ROB & 0.438 & \textbf{0.310} &  0.322 & 0.516 & 0.514 & \textbf{0.511}  & 0.313 & 0.265 & \textbf{0.239} & 0.656  & 0.563 & \textbf{0.159} \\
ECE & 0.345 & \textbf{0.163} &  0.232 & 0.493 & \textbf{0.436} & 0.447  & 0.243  & 0.278 & \textbf{0.206} & 0.658  & 0.534 & \textbf{0.118} \\
\hline
& \multicolumn{3}{c|}{RACE} & \multicolumn{3}{c|}{Twitter} & \multicolumn{3}{c|}{RACE} & \multicolumn{3}{c}{Twitter} \\
\cline{2-13}
& Vanilla & QA & KT & Vanilla & QA & KT & Vanilla & QA & KT & Vanilla & QA & KT \\
\hline
ACC & 0.538 & 0.549 & \textbf{0.630} & 0.662 & \textbf{0.718} & 0.716  & 0.561 & 0.654 & \textbf{0.670} & 0.589  & 0.728 & \textbf{0.743} \\
ROB & 0.423 & 0.440 & \textbf{0.345} & 0.297 & \textbf{0.244} & 0.253  & 0.417 & 0.272 & \textbf{0.249} & 0.411 & 0.269 & \textbf{0.115} \\
ECE & 0.343 & 0.274 & \textbf{0.273} & 0.245 & 0.284 & \textbf{0.173}  & 0.352 & 0.278 & \textbf{0.232} & 0.358 & 0.206 & \textbf{0.087} \\
\hline
& \multicolumn{3}{c|}{ARC} & \multicolumn{3}{c|}{GooglePlay} & \multicolumn{3}{c|}{ARC} & \multicolumn{3}{c}{GooglePlay} \\
\cline{2-13}
& Vanilla & QA & KT & Vanilla & QA & KT & Vanilla & QA & KT & Vanilla & QA & KT \\
\hline
ACC & 0.427 & 0.619 & \textbf{0.636} & 0.740 & 0.787 & \textbf{0.824}  & \textbf{0.729} & 0.648 & 0.694 & 0.722  & 0.804 & \textbf{0.829} \\
ROB & 0.561 & 0.381 &  \textbf{0.315} & 0.221 & \textbf{0.120} & 0.139  & \textbf{0.256} & 0.336 & 0.270 & 0.252 & 0.186 & \textbf{0.140} \\
ECE & 0.473 & 0.275 &  \textbf{0.240} & 0.143 & 0.139 & \textbf{0.039}  & \textbf{0.210} & 0.327 & 0.226 & 0.202  & 0.174 & \textbf{0.082} \\
\hline

\end{tabularx}
\end{table*}

\subsection{Quantity Optimization of CoT for Efficient Training}
\begin{figure}[h]
\centering
\begin{subfigure}[b]{0.23\textwidth}
  \includegraphics[width=\textwidth]{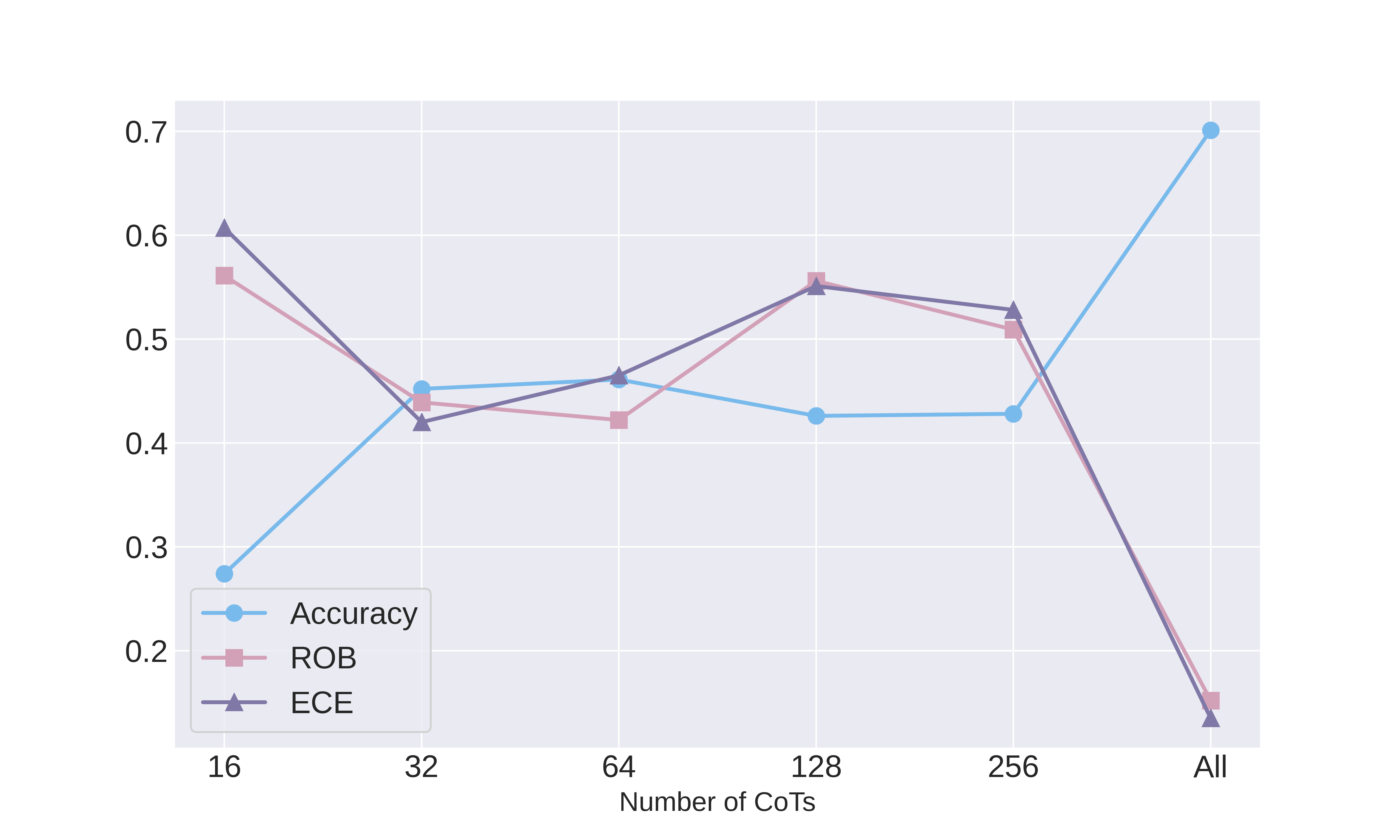}
  \caption{Truthfulqa}
  \label{fig:sub1}
\end{subfigure}
\hfill
\begin{subfigure}[b]{0.23\textwidth}
  \includegraphics[width=\textwidth]{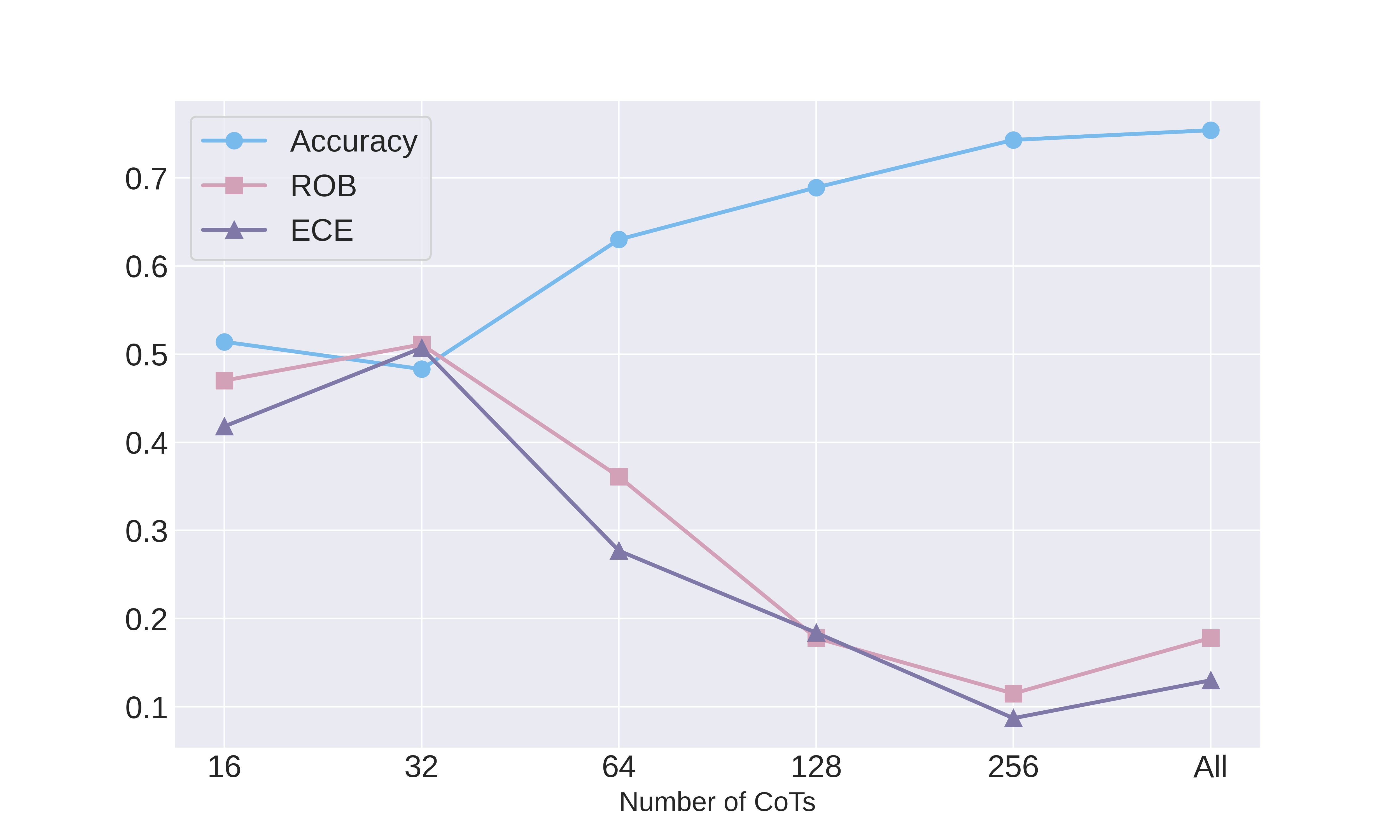}
  \caption{Twitter}
  \label{fig:sub2}
\end{subfigure}

\begin{subfigure}[b]{0.23\textwidth}
  \includegraphics[width=\textwidth]{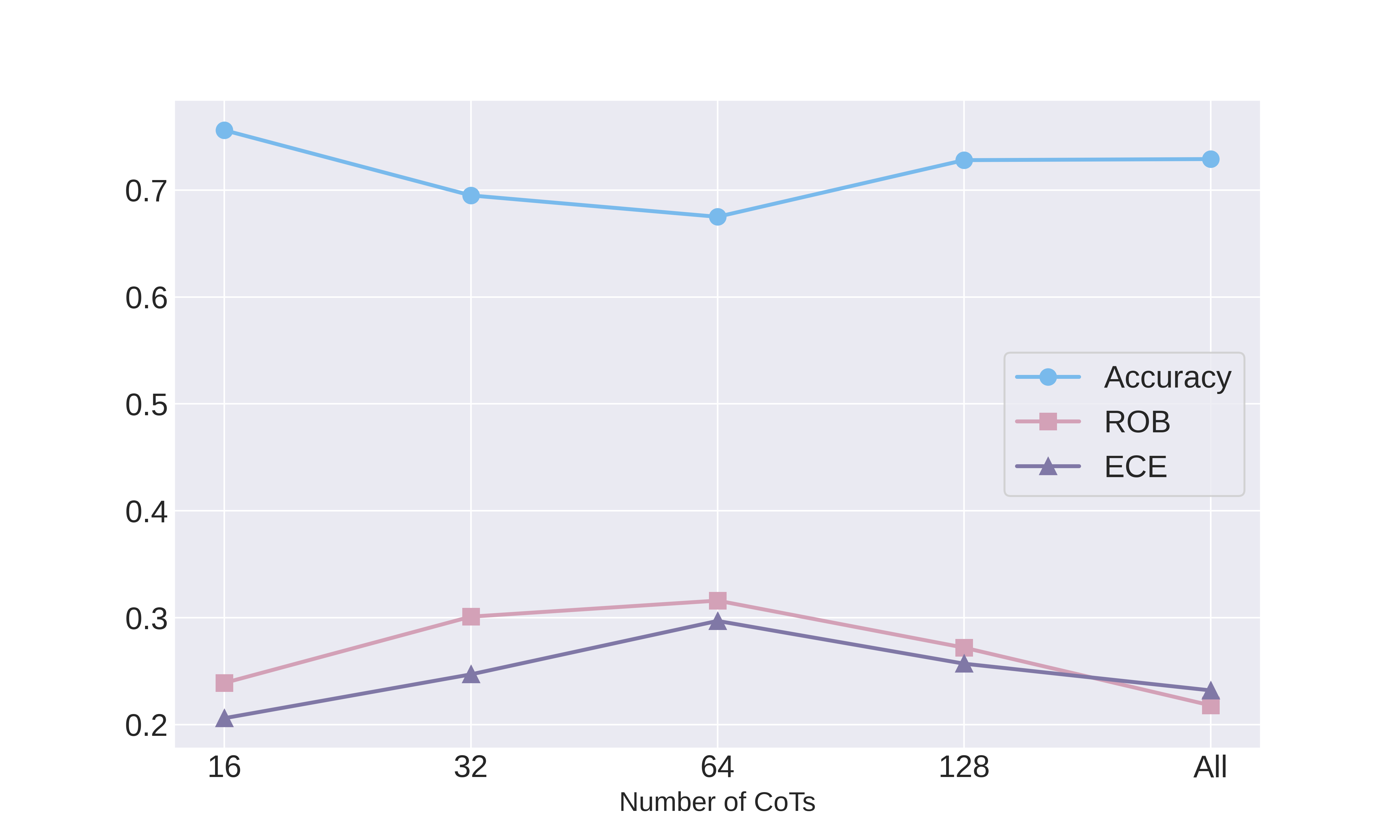}
  \caption{McTest}
  \label{fig:sub3}
\end{subfigure}
\hfill
\begin{subfigure}[b]{0.23\textwidth}
  \includegraphics[width=\textwidth]{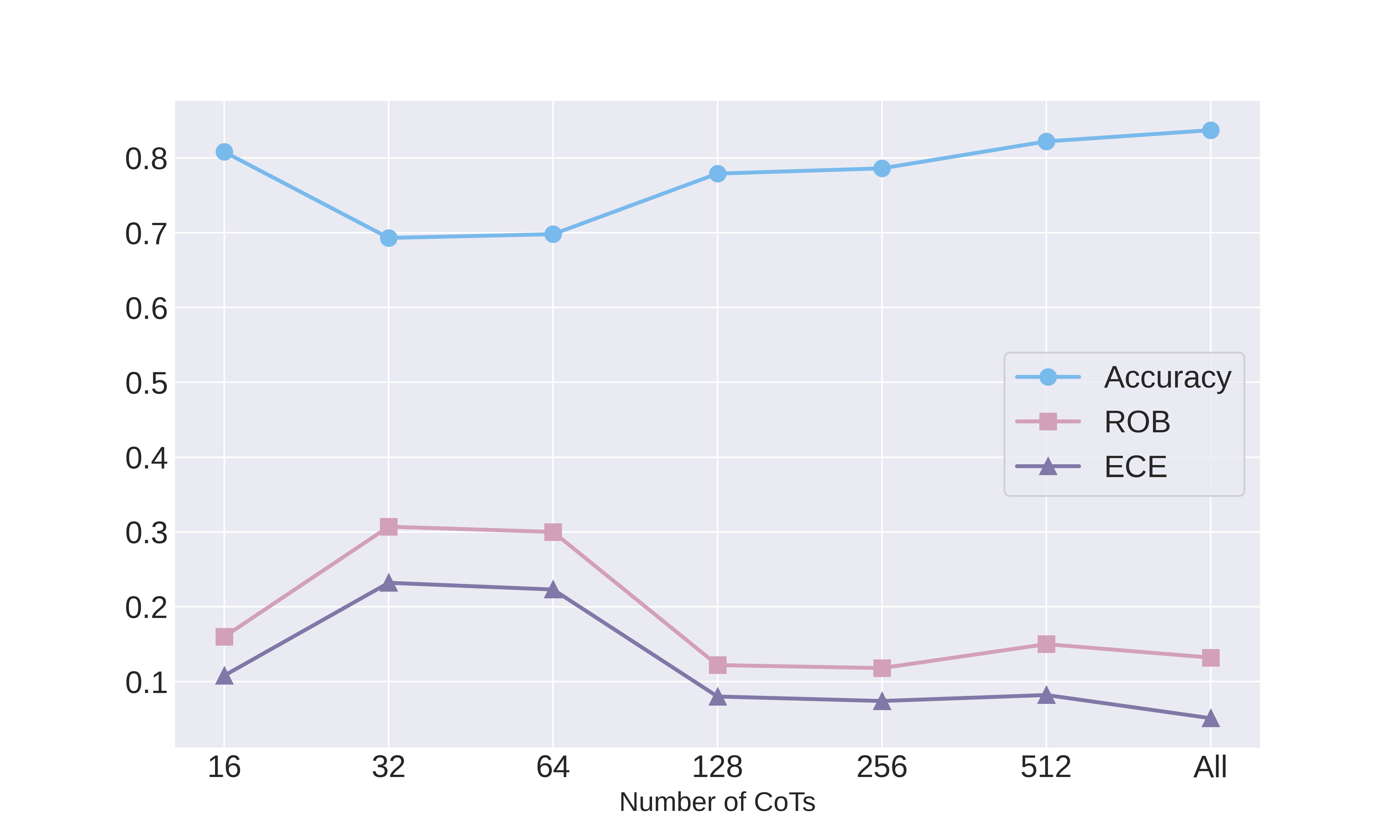}
  \caption{SST2}
  \label{fig:sub4}
\end{subfigure}

\caption{Performance trends in fine-tuning the Vicuna-7B with different quantities of CoT on four datasets.}
\label{fig3}
\end{figure}

We further examined the impact of CoT quantity on model performance using Vicuna-7B as a case study. As illustrated in Figures \ref{fig:sub1} and \ref{fig:sub2}, the performance improved as the number of CoTs increases, being consistent with our intuition. However, as shown in Figures \ref{fig:sub3} and \ref{fig:sub4}, we made a surprising discovery that a few quantity of CoTs, such as 16, can achieve a decent performance, demonstrating the potential of the KT method for cost-effective training. This finding provides guidance for the quantity optimization of CoT.

\section{Analysis}

\subsection{Key Reasons Behind the Effectiveness of KT Method}
The primary cause of overconfidence bias in models stems from their limited parameters, leading to a deficiency in the knowledge for accurately answering questions, which often results in arbitrary responses. Through KT method, smaller models can assimilate the logic and knowledge from larger models. By selecting only those CoTs that correctly answer questions as training data, smaller models can better understand and judge these questions. Consequently, this leads to more precise outputs from these models, matched with accurate confidence levels.

\subsection{Analysis of Comparative Results}
\begin{figure}[h]%
    \centering
    \small
    \begin{subfigure}[t]{0.23\textwidth} % Adjust the width to fit side by side
        \includegraphics[width=\textwidth]{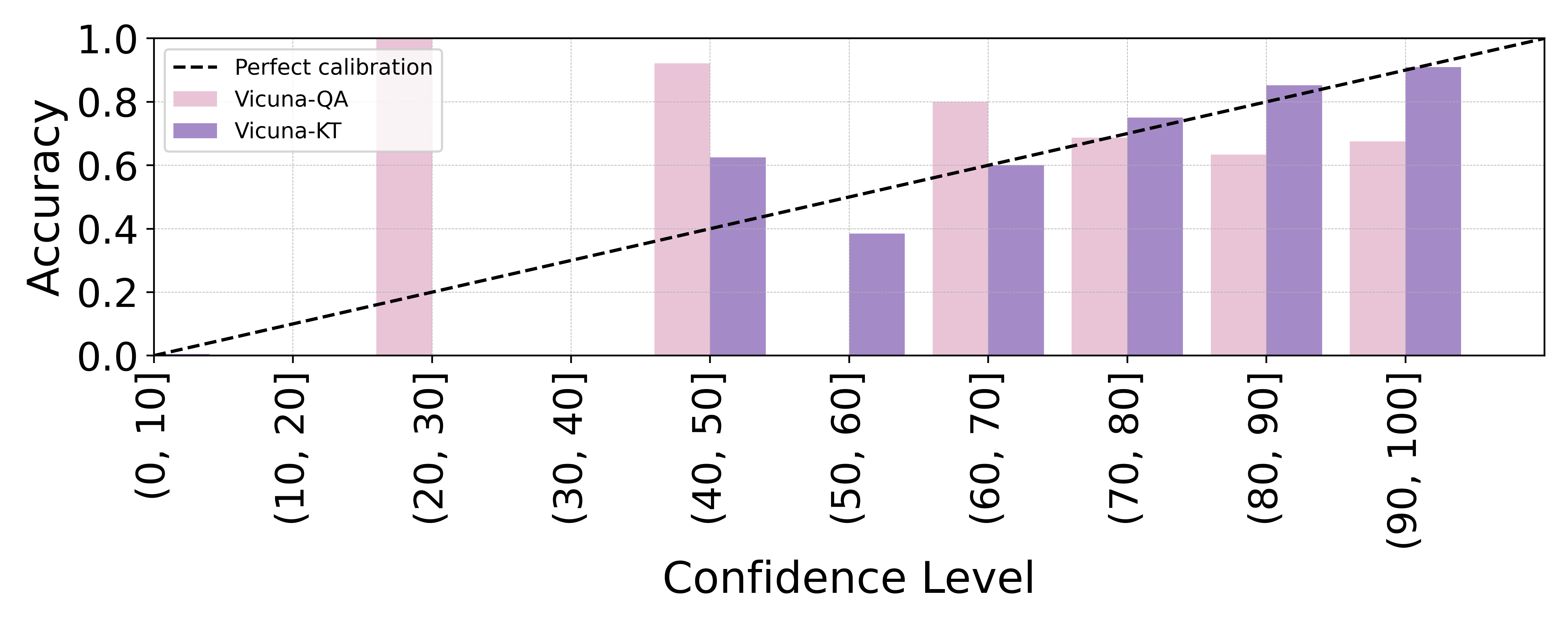}
        \caption{SST2}
        \label{fig4(a)}
    \end{subfigure}
    \hfill % This will add some space between the two subfigures
    \begin{subfigure}[t]{0.23\textwidth} % Adjust the width here as well
        \includegraphics[width=\textwidth]{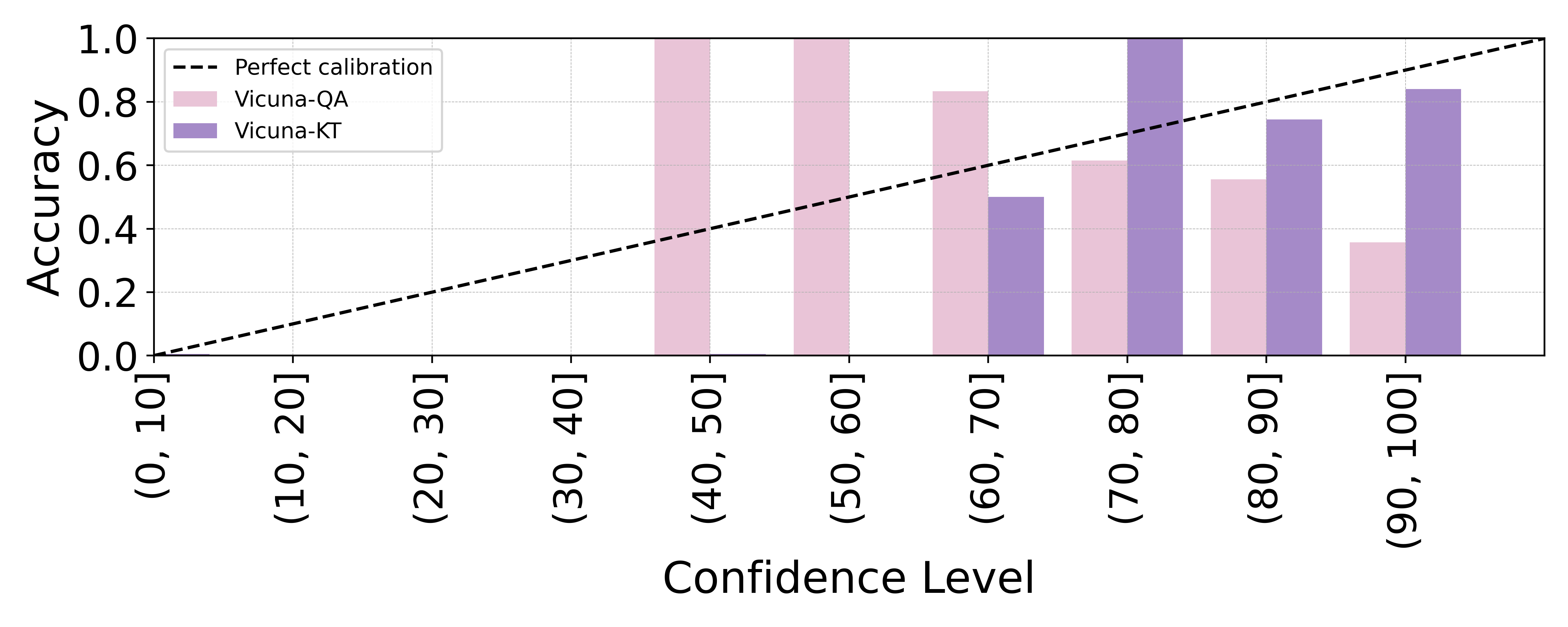}
        \caption{Financial Phrasebank}
        \label{fig4(b)}
    \end{subfigure}
    \caption{Calibration curves comparing model accuracy and confidence levels across two datasets in Vicuna.}
    \label{fig4}
\end{figure}

In Table \ref{table3}, we have validated the effectiveness of the KT method. As shown in Figure \ref{fig4}, the calibration curves for two datasets demonstrate KT's enhancement in model reliability. Graphs for other datasets are presented in the Appendix \ref{appendix4}. However, we observed that for the ARC dataset with Vicuna baseline, Vanilla model overperformed both QA and KT fine-tuning methods, a result not replicated on LLaMA. We attribute this to Vicuna's original fine-tuning on LLaMA, which gave it superior zero-shot capabilities. Combined with the unique characteristics of the dataset, this occasionally leads to such anomalies.

\section{Conclusion}
The successful implementation of CoT knowledge transfer has enabled “small” LLMs to effectively assimilate knowledge from “big” LLMs. This has significantly reduced their inherent overconfidence bias due to limited parameters, thereby enhancing their reliability and applicability in various tasks. 

\section*{Limitations}
The knowledge transfer method we propose, due to its reliance on fine-tuning the model through CoT, occasionally results in the generation of longer tokens. Sometimes, the model may fall into self-dialogue. Additionally, for classification tasks, tokens other than the labels can sometimes be redundant and unnecessary.

\bibliography{custom}
\bibliographystyle{acl_natbib}

\newpage
\onecolumn
\appendix

\section{Example of prompt to inference and corresponding output}
\label{appendix1}
\begin{promptbox}
\noindent \textbf{Prompt:} Read this sentence, select the correct sentiment for it and give the option letter: A: positive, B: negative. Use the following format to provide your answer and confidence level.\\ 
\noindent \textit{Answer and Confidence (0-100):} [Your answer, please only include the capital letter], [Your confidence level, please only include the numerical number]\% 

\noindent \textit{Note:} The confidence level indicates the degree of certainty you have about your answer and is represented as a percentage. For instance, if your confidence level is 80\%, it means you are 80\% certain that your answer is correct and there is a 20\% chance that it may be incorrect.

\noindent \textit{Sentence:} [it shows us a slice of life that's very different from our own and yet instantly recognizable]

\noindent \textbf{Answer and Confidence (0-100):} A, 90

\end{promptbox}

\section{The details of each dataset}
\label{appendix2} 

\begin{table}[h]
\centering
\small
\caption{Overview of Five Multi-choice Questions Datasets}
\begin{tabular}{p{3cm}p{6cm}p{3cm}}
\toprule
\textbf{Dataset} & \textbf{Description} & \textbf{Construction} \\
\midrule
Truthfulqa & Truthfulqa is a dataset that focuses on assessing the truthfulness of answers, where models are evaluated based on their ability to provide accurate and factually correct responses to questions. & 817 questions in total. \\
McTest & The McTest dataset contains 660 fictional stories, with each story having 4 questions and 4 candidate answers. & There are 2000 questions. \\
RACE & The RACE dataset is a large-scale reading comprehension dataset collected from English examinations in middle and high schools. & 87866 in training set, 4887 in validation set, 4934 in testing set.\\
ARC & ARC dataset is an English question-answering exam dataset with 7787 elementary-level multiple-choice, divided into 2590 difficult and 5197 simple questions. & 3370 in training set, 869 in validation set, 3548 in testing set.\\
\bottomrule
\end{tabular}

\end{table}

\begin{table}[h]
\centering
\small
\caption{Overview of Four Sentiment Analysis Datasets.}
\begin{tabular}{p{3cm}p{6cm}p{3cm}}
\toprule
\textbf{Dataset} & \textbf{Description} & \textbf{Construction}\\
\midrule
SST-2  & SST-2 is a dataset for sentiment analysis, containing movie review sentences labeled as positive or negative. & 67349 in training set, 872 in validation set, 1821 in testing set. \\
Financial Phrasebank & Financial Phrasebank is a dataset tailored for financial news sentiment analysis, containing sentences from financial news articles labeled with sentiment categories such as positive, negative, or neutral. & 14780 data in total. \\
Twitter & Twitter Sentiment Analysis contains 1,578,627 classified tweets, each row is marked as 1 for positive sentiment and 0 for negative sentiment. & 119988 in training set, 29997 in validation set, 61998 in testing set. \\
GooglePlay & GooglePlay is a dataset containing players' reviews on games, labeled as positive or negative. & 108837 comments in total.\\
\bottomrule
\end{tabular}
\end{table}

\newpage

\section{Performance trends in fine-tuning the Vicuna-7B with different quantities of CoT on other four datasets.}
\label{appendix3}

\begin{figure*}[h]
    \small
    \centering

    \begin{subfigure}{0.23\textwidth}
        \includegraphics[width=\textwidth]{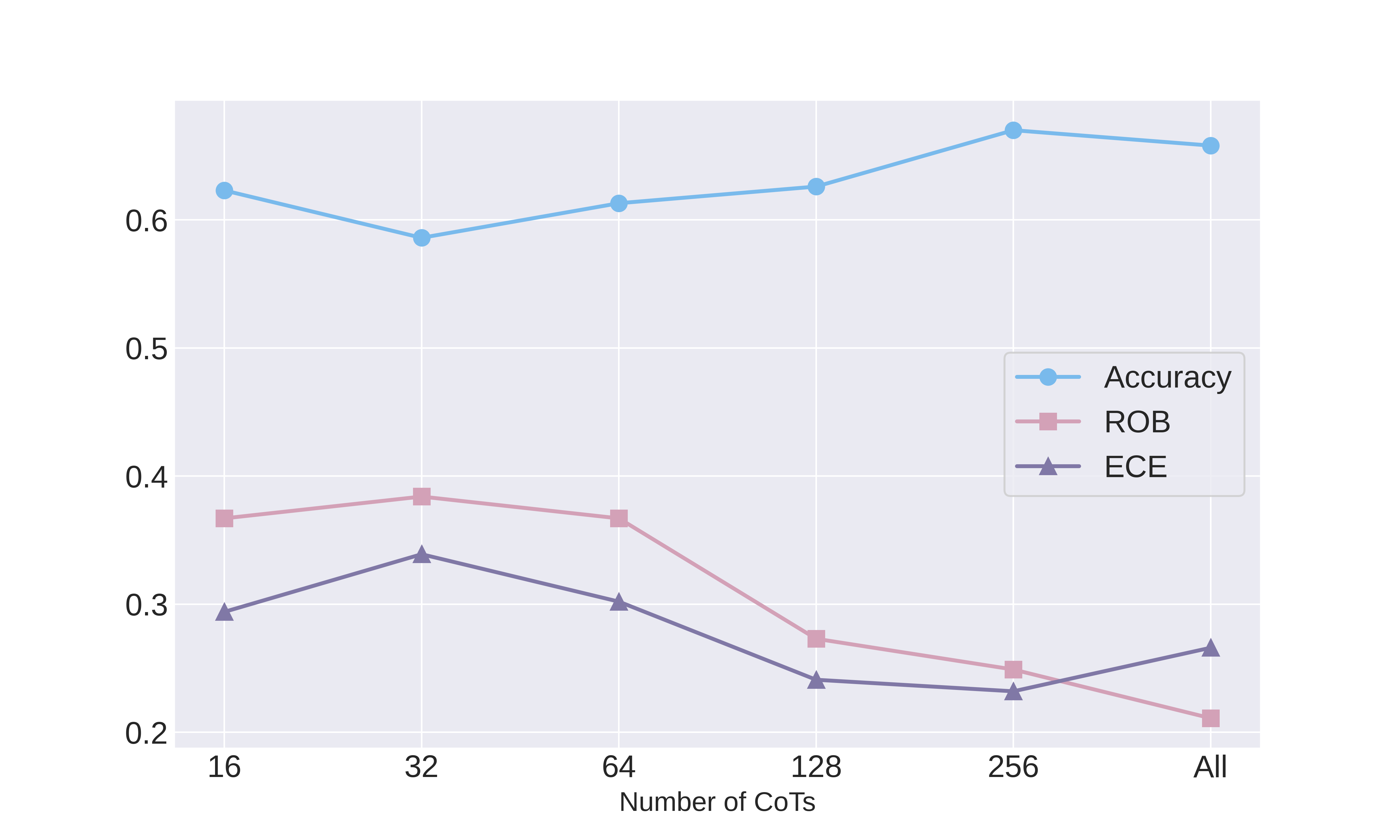}
        \caption{RACE}
    \end{subfigure}\hfill
    \begin{subfigure}{0.23\textwidth}
        \includegraphics[width=\textwidth]{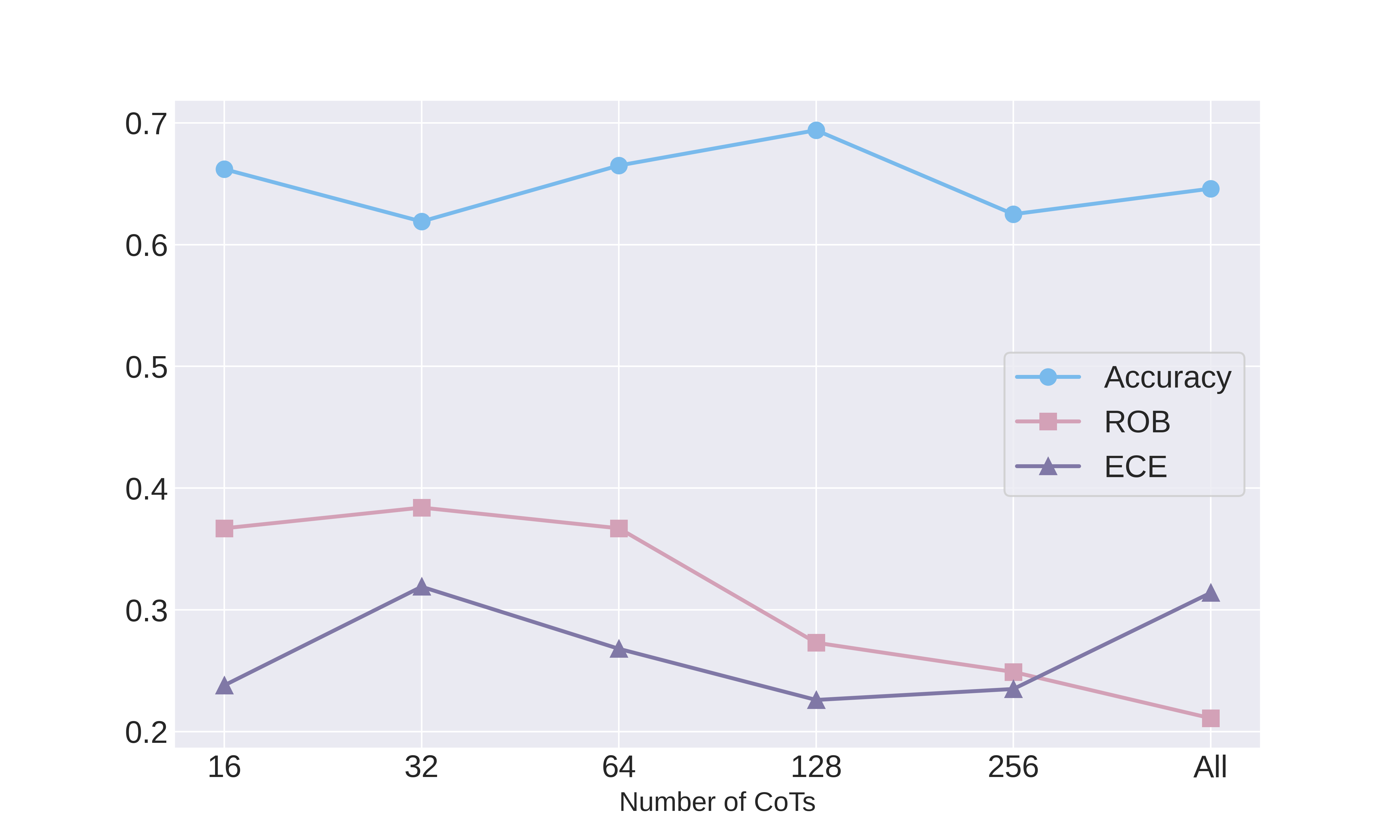}
        \caption{ARC}
    \end{subfigure}
    \hfill
    \begin{subfigure}{0.23\textwidth}
        \includegraphics[width=\textwidth]{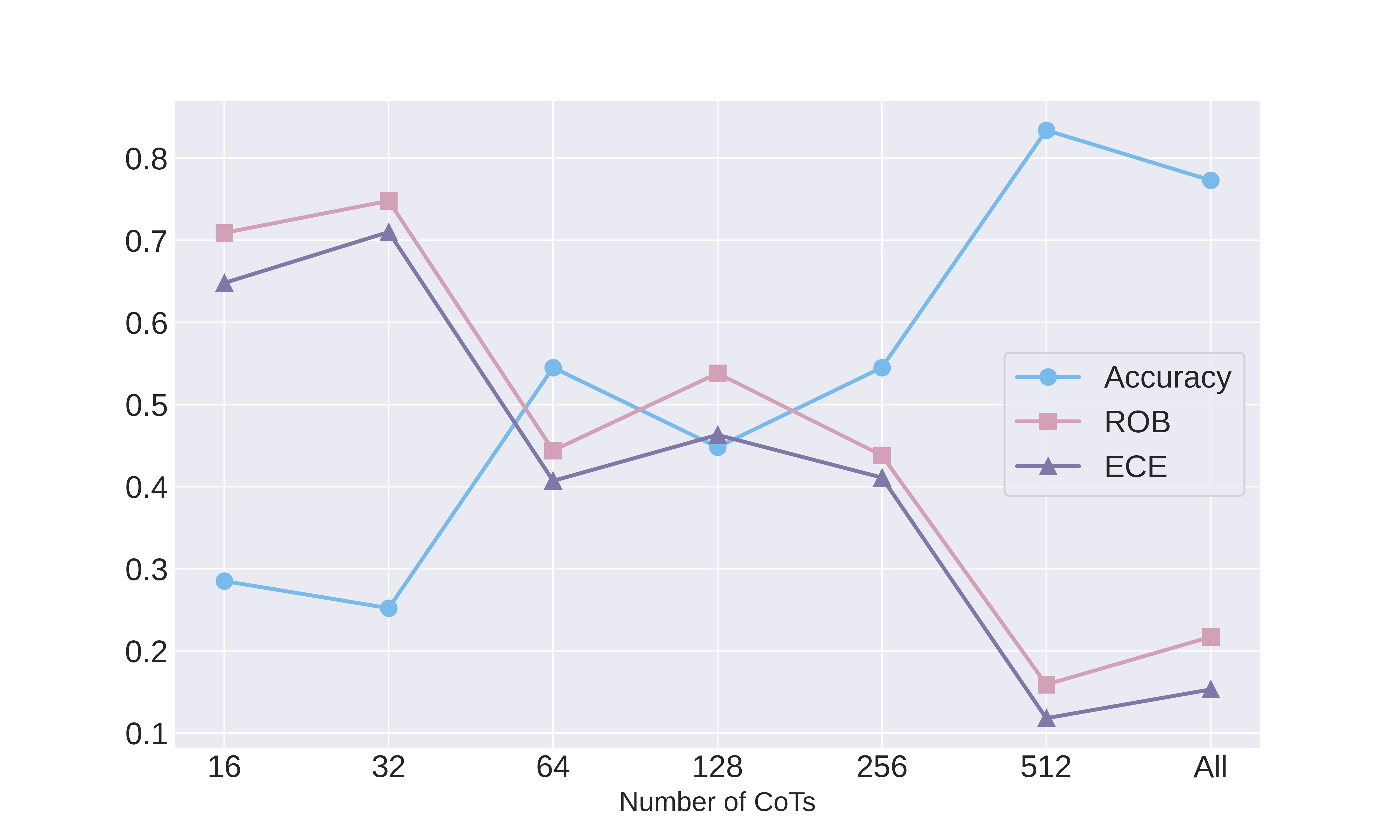}
        \caption{Financial Phrasebank}
    \end{subfigure}\hfill
    \begin{subfigure}{0.23\textwidth}
        \includegraphics[width=\textwidth]{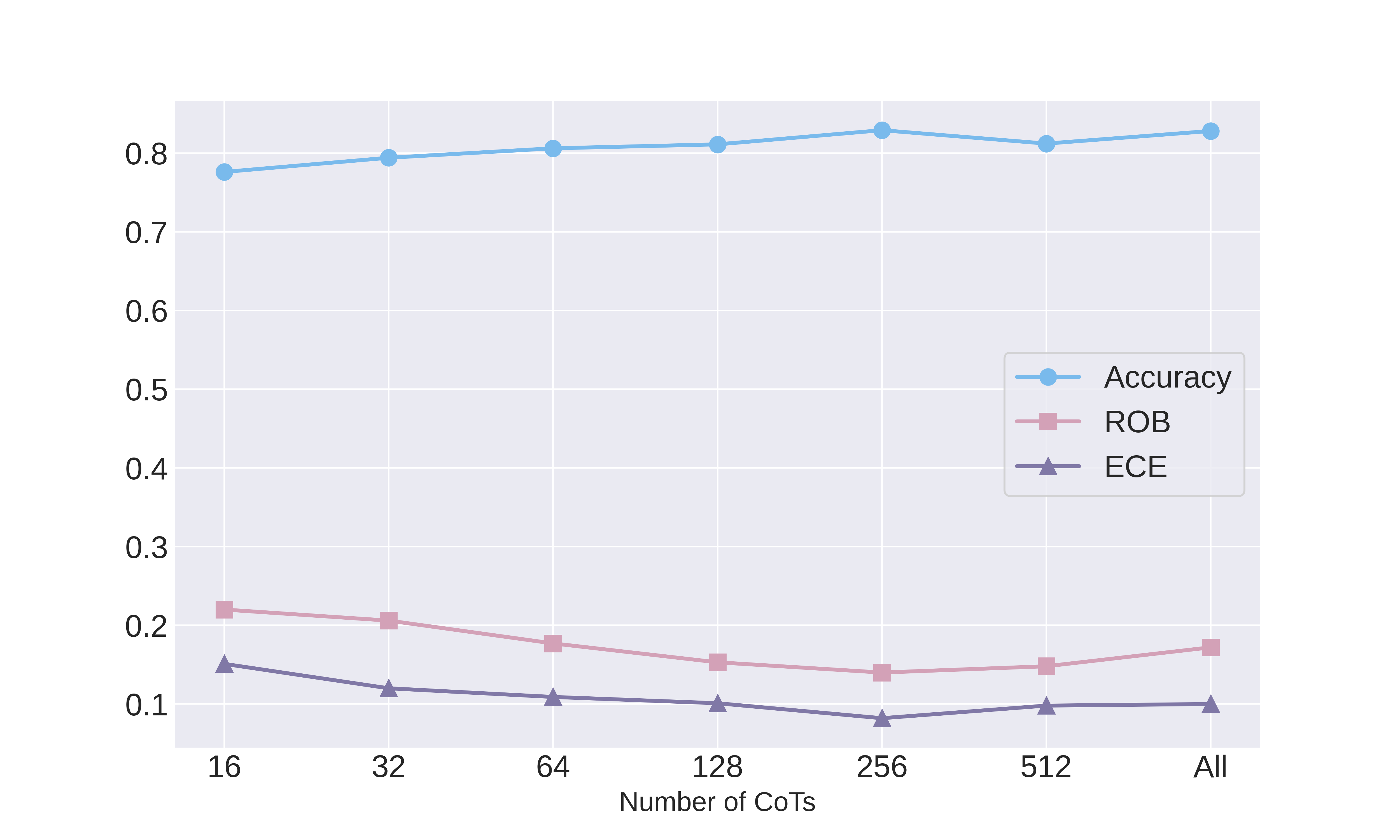}
        \caption{GooglePlay}
    \end{subfigure}
\end{figure*}

\section{Calibration curves comparing model accuracy and confidence levels across other datasets in Vicuna}
\label{appendix4}

\begin{figure}[h]%
    \centering
    \small
    \begin{subfigure}[t]{0.45\textwidth} % Adjust the width to fit side by side
        \includegraphics[width=\textwidth]{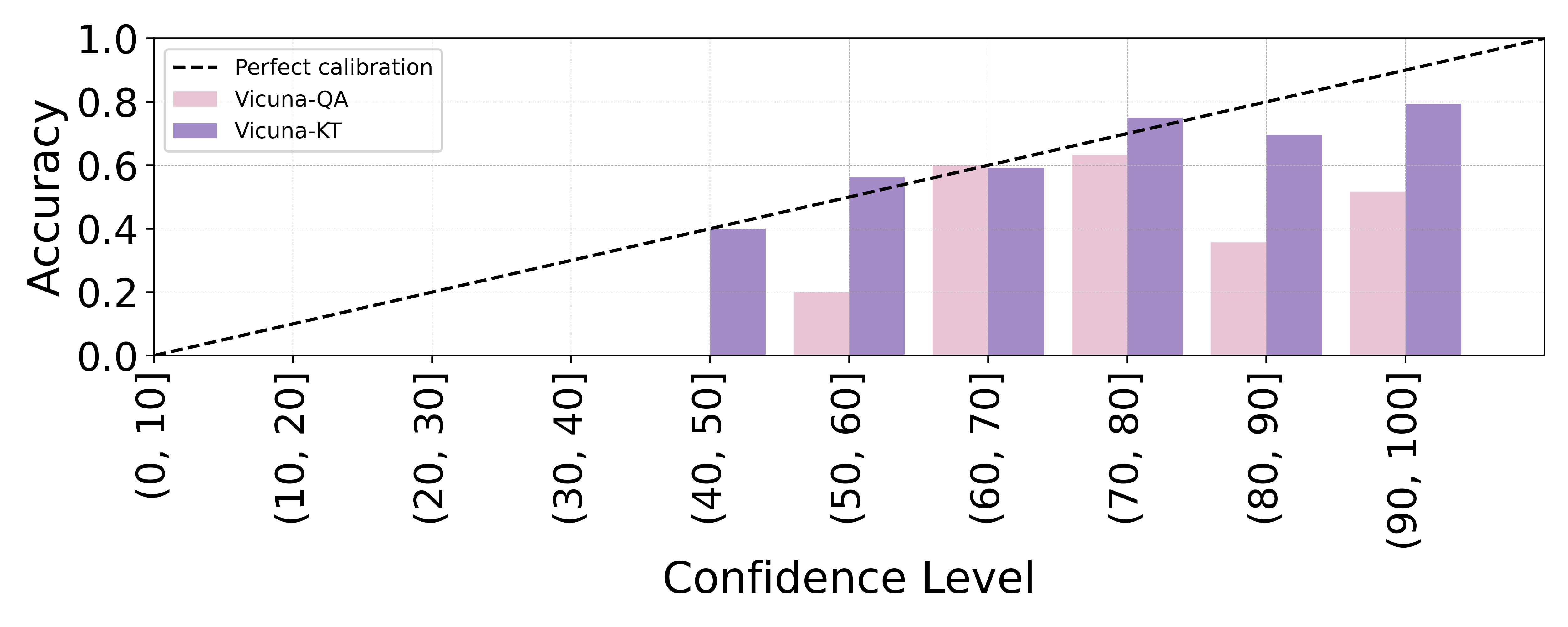}
        \caption{Truthfulqa}
    \end{subfigure}
    \hfill % This will add some space between the two subfigures
    \begin{subfigure}[t]{0.45\textwidth} % Adjust the width here as well
        \includegraphics[width=\textwidth]{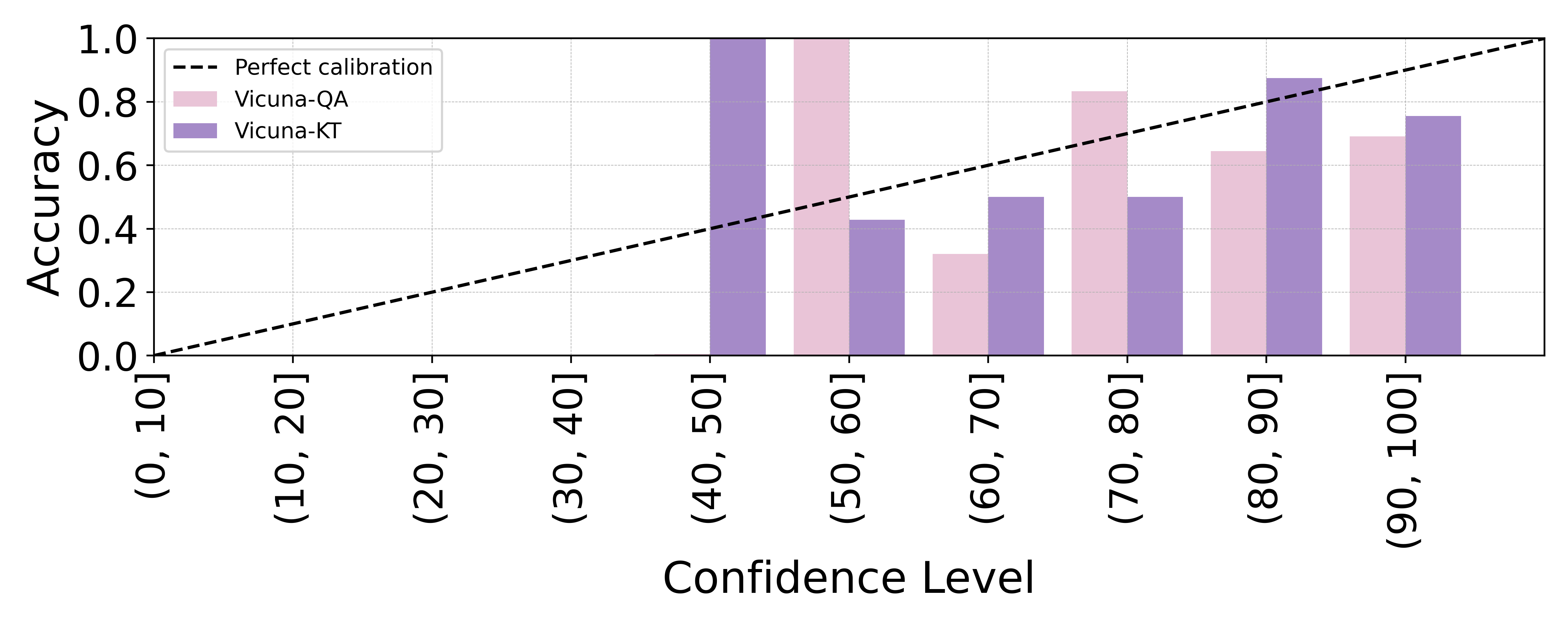}
        \caption{McTest}
    \end{subfigure}
    \hfill % This will add some space between the two subfigures
    \begin{subfigure}[t]{0.45\textwidth} % Adjust the width here as well
        \includegraphics[width=\textwidth]{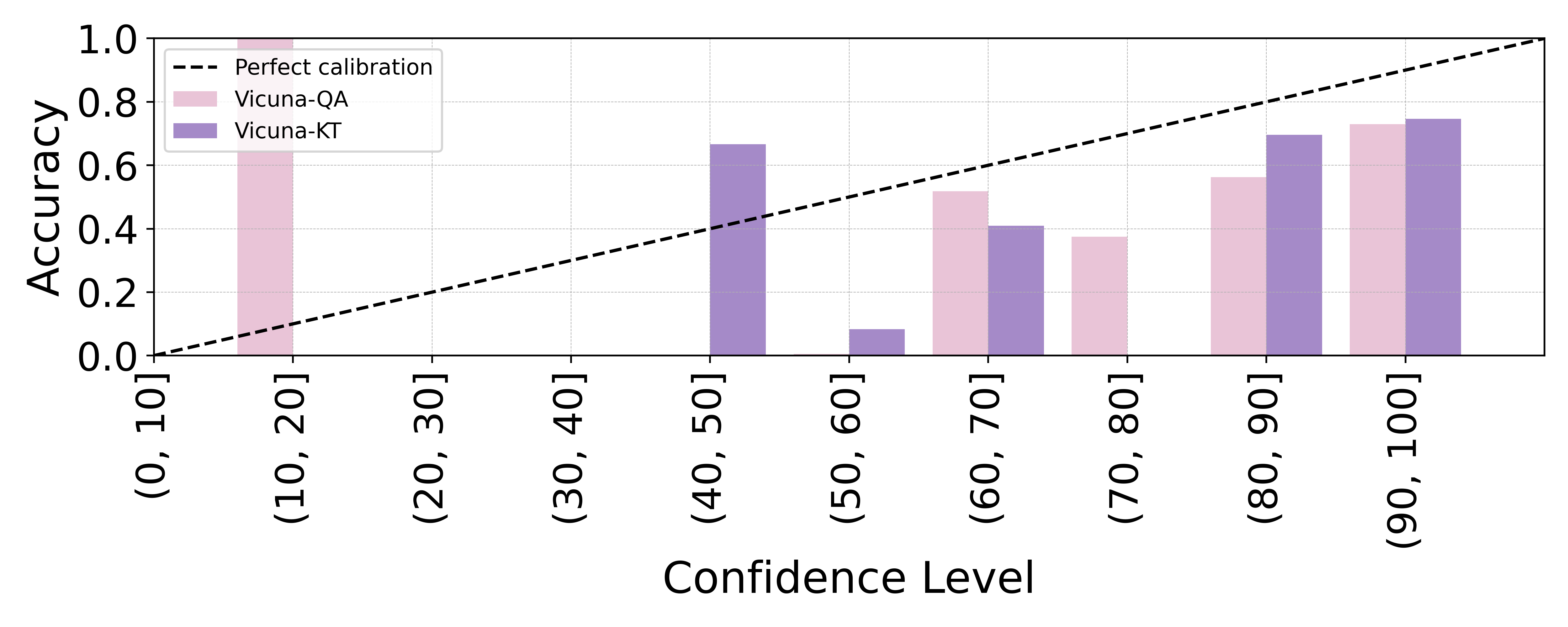}
        \caption{RACE}
    \end{subfigure}
    \hfill % This will add some space between the two subfigures
    \begin{subfigure}[t]{0.45\textwidth} % Adjust the width here as well
        \includegraphics[width=\textwidth]{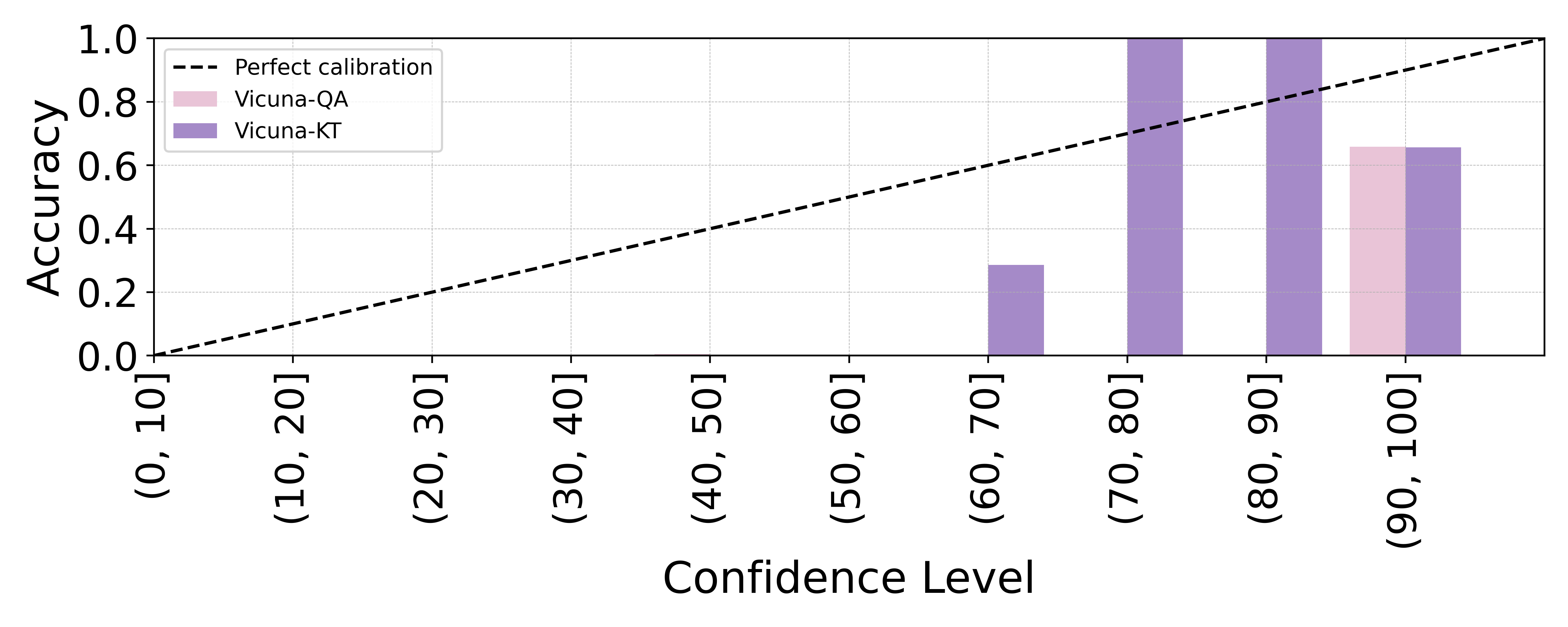}
        \caption{ARC}
    \end{subfigure}
    \hfill % This will add some space between the two subfigures
    \begin{subfigure}[t]{0.45\textwidth} % Adjust the width here as well
        \includegraphics[width=\textwidth]{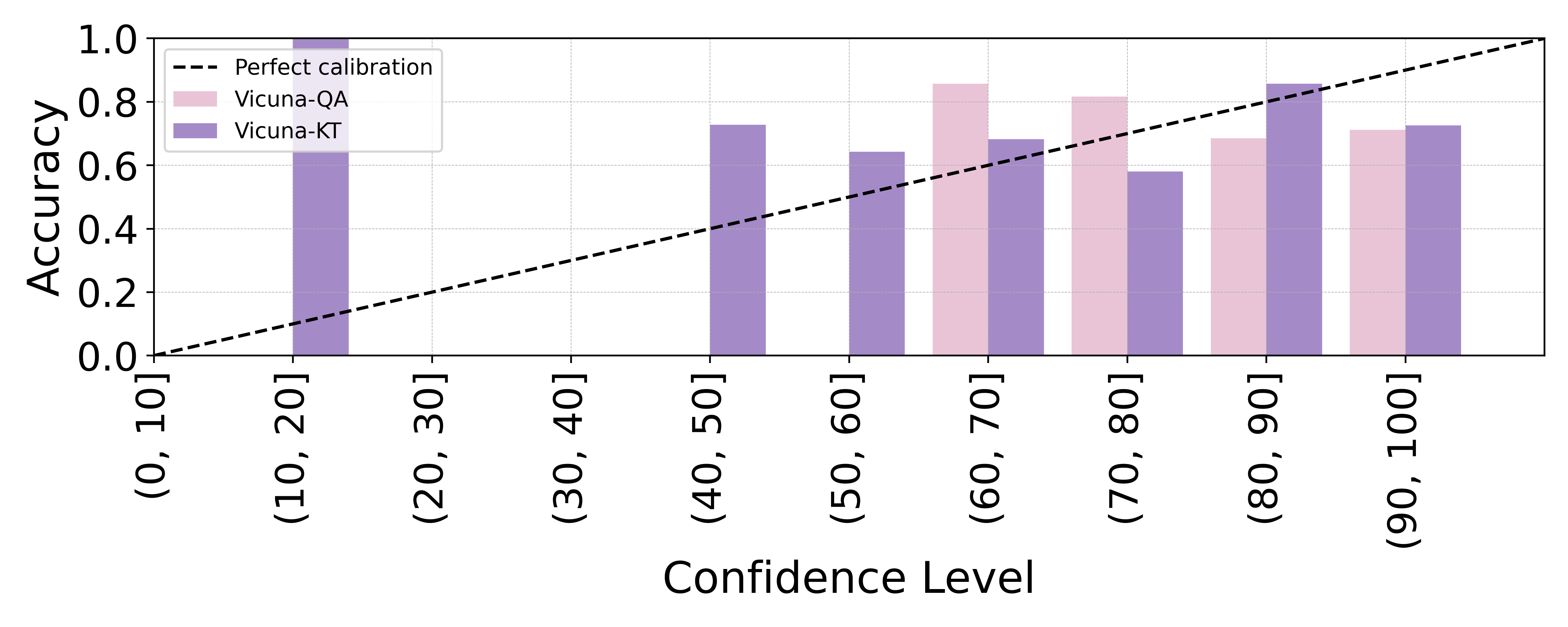}
        \caption{Twitter}
    \end{subfigure}
    \hfill
    \begin{subfigure}[t]{0.45\textwidth} % Adjust the width here as well
        \includegraphics[width=\textwidth]{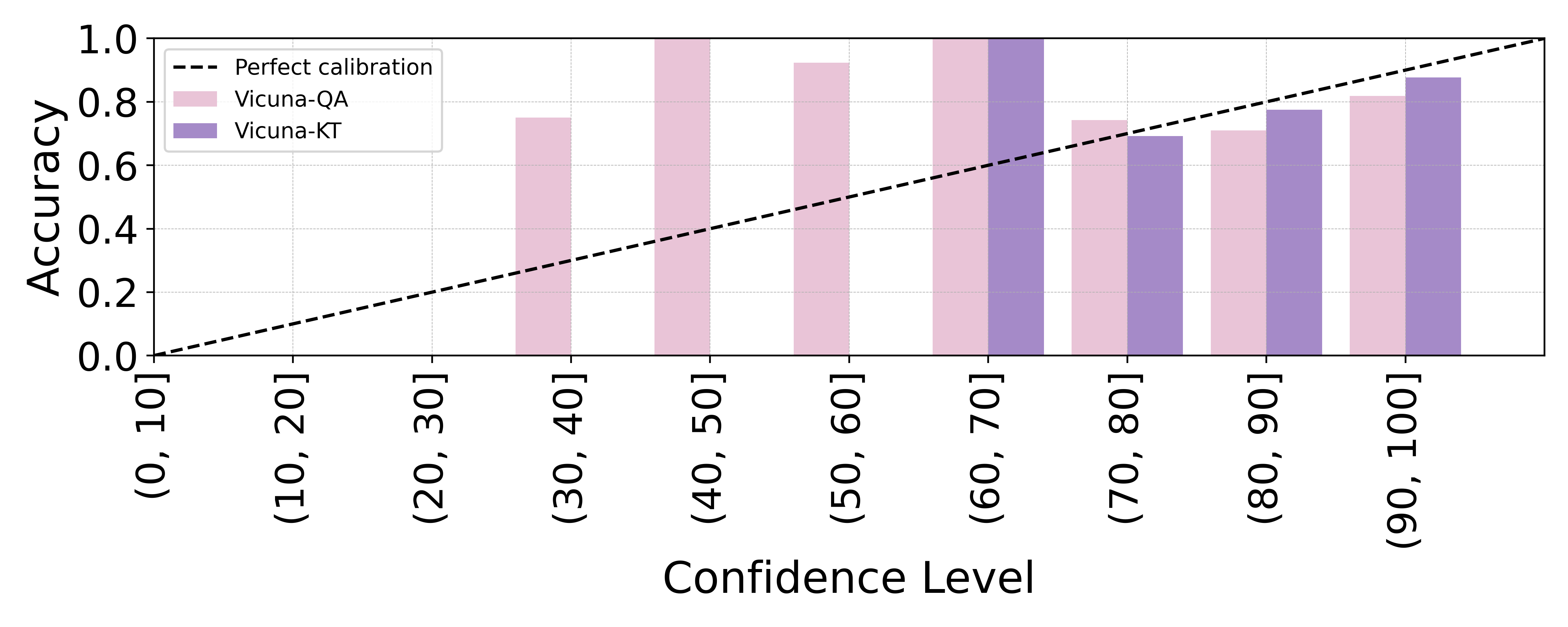}
        \caption{GooglePlay}
    \end{subfigure}
\end{figure}

\end{document}